# HMM Speaker Identification Using Linear and Non-linear Merging Techniques

*Unathi Mahola, Fulufhelo V. Nelwamondo, Tshilidzi Marwala*

School of Electrical and information Engineering
University of the Witwatersrand,
Johannesburg, South Africa
ratyie2002@yahoo.com    nelwamof@undergrad.ee.wits.ac.za    t.marwala@ee.wits.ac.za

## Abstract

Speaker identification is a powerful, non-invasive and inexpensive biometric technique. The recognition accuracy, however, deteriorates when noise levels affect a specific band of frequency. In this paper, we present a sub-band based speaker identification that intends to improve the live testing performance. Each frequency sub-band is processed and classified independently. We also compare the linear and non-linear merging techniques for the sub-bands recognizer. Support vector machines and Gaussian Mixture models are the non-linear merging techniques that are investigated. Results showed that the sub-band based method used with linear merging techniques enormously improved the performance of the speaker identification over the performance of wide-band recognizers when tested live. A live testing improvement of 9.78% was achieved.

## 1. Introduction

Speaker identification and speaker verification form a larger discipline of speech recognition. Speaker Identification (SID) tries to determine which speaker generated a speech signal whereas speaker verification confirms if the portion of the speech belongs to the individual who claims it. It should be noted that there are two types of speaker identification, which are; text dependent and text independent. This paper will however focus on text dependent speaker identification.

Current text dependent or even independent SID systems produce reasonable results, but still lack the necessary performance if they are to be used by the general public (i.e. live testing). This paper challenges this problem by extending the idea originally presented by Fletcher [1]. Fletcher's work resulted to the introduction of the sub-bands analysis of speech signals. This method shows an advantage in speech recognition and will therefore be further explored.

In exploring this topic, first, a brief review of the work done in sub-band based SID is evaluated. Thereafter relevant mathematical background is briefly presented. Both the conventional and the slightly improved sub-band based speaker identification are also discussed. A reliable confidence measure is essential in SID and for this reason this is also discussed. Finally, the results and conclusion of the sub-band based speaker identification using the conventional Hidden Markov Model (HMM) method as a baseline are presented.

## 2. Related work and motivation

Fletcher [1] suggested that linguistic messages in humans get decoded independently in different frequency bands and the final decision is based on combining decisions from the sub-bands. His work resulted to the beginning of the sub-bands analysis of speech signals. Research in this area proved that this method outperforms the classical wide band speaker identification systems.

Sub-band approach has now become popular in speech recognition. Some researchers who have shown a lot of interest in this field are Hermansky, Mirghafori and Damper [2-4]. All these researchers' results revealed that the sub-band process is highly dependent on the merging technique used. Various merging technique have been used, for instance, Mirghafori and Hermansky compared MLP with linear based merger. A number of other merging techniques have been proposed but no comparative work has been performed using SVM or GMM merger. Due to this, the authors will perform comparative work between SVM, GMM and linear based mergers. Even though both the conventional and sub-band based results give reasonable results, they still lack reliable performance if they are live tested. Given this problem, the paper also attempts to develop an approach that will improve live testing by using both conventional and sub-band based SID recognizers.

It should be noted that the proposed sub-band based identification system uses a number of recognizers and therefore it can be computationally intensive to perform exhaustive search to find the optimum architecture for each band. As a result Genetic Algorithm (GA) optimization is used to find the optimum architectures of all recognizers.

## 3. Background

Based on the discussion present above, this section presents a brief discussion of HMM, GMM, SVM and GA, as they will be used through out this paper.

### 3.1 Hidden Markov Model (HMM)

HMM recognisers are used by both the conventional or baseline approach and the sub-band based recognizer. The HMM model that will be used is the left-to-right or barks models. This model is mostly used in speech or speaker recognition. The probability of speech feature vectors generated from an HMM is computed using the transition probabilities between states and the observation probabilities of feature vectors in a given state.

There are three central HMM problems in finding the probability of speech feature vectors generated from an HMM [5]. Firstly, evaluation, which finds the probability that a sequence of visible states was generated by the model *M* and this, is solved by the Forward and Viterbi algorithms [5]. Secondly, decoding finds state sequence that maximizes probability of observation sequence using Viterbi algorithm. Lastly, training which adjusts model parameters to maximize probability of observed sequence. This last step is simply a problem of determining the reference speaker model for all speakers. The Baum-Welch re-estimation procedures or Forward-Backwards Algorithm are used for this case as presented in [5].

There are two known models of the HMM which are: Continuous HMM (CHMM) and Discreet HMM (DHMM). Research reveals that data is lost when modelling using DHMM [5], since output probabilities are computed using quantised codebook and for this reason only CHMM will be investigated.

### 3.2 Gaussian Mixture Models (GMM)

The GMM recognizer is used by the sub-band based SID system as a linear merger. Using GMM, the speaker identity is computed by finding the likelihoods of the unknown speaker utterance given the speaker models [6]. These log-likelihoods are in turn used to estimate the confidence of the system and are given by,

$$L_s(X) = \log P(X | \lambda_s) \quad (1)$$

The speaker models are estimated using the EM algorithm. For the input vector $X$ the mixture densities of the speaker, $S$, is computed as [6],

$$P(X | \lambda_s) = \sum_{i=1}^{M} P_i^s b_i^s(x) \quad (2)$$

where, $P_i^s$ are the mixture weights and $b_i^s(x)$ are the Gaussian densities which are parameterized by a mean vector and covariance matrix.

### 3.3 Support Vector Machines (SVM)

SVM works by creating the linear hyper-plane function and maximizes the margin of this function [7]. To illustrate this, assume the training set is $(x_i, y_i)$. SVM assumes this set is linearly separable, meaning there exist [$w, b$] so that,

$$y_i(X_i^T w + b) - 1 \geq 0 \quad (3)$$

It should be noted that when the data is not linearly separable, a slack variable, $\xi_i$, is introduced and the optimization problem becomes [7],

$$\min_{w, b, \xi} \quad \|w\|^2 + C \sum_{i=1}^{l} \xi_i \quad (4)$$

Equation 4 is solved using quadratic programming by finding the Lagrangian multiplier and applying the Karush-Kuhn Tucker (KKT) conditions. The SVM classifier described in this section is utilized by the merger of the sub-band based SID and this is discussed further in section 4.

### 3.4 Genetic Algorithm (GA)

GA is a non deterministic optimization technique that uses the concept of survival of the fittest over consecutive generations to solve optimization problems. The reader must note that this optimization method is used to find the optimum architecture of HMM and GMM. GA is also used to find the optimum confidence log-likelihood ratio. The GA algorithm implemented uses a population of string chromosomes, which represent a point in the search space. This algorithm is implemented by three main procedures which are selection process, crossover and mutation. More detail on GA is presented by [8].

## 4. Proposed methodology

### 4.1 Pre-processing and feature extraction

The first essential step in any speech related systems is speech pre-processing and feature extraction. Starting with the pre-processing stage, this stage can be sub-divided into various stages as shown in figure 1.The first stage of the pre-processing is to apply the pre-emphasis filter. This filter ensures that the effect of clipping is reduced on the input signal. The next stage of this pre-processing stage is to isolate the utterance from the background noise by detecting the beginning and the end of the utterance. The speech signal is further blocked into frames and a hamming window is applied to each frame to avoid signal discontinuities at the beginning of each frame.

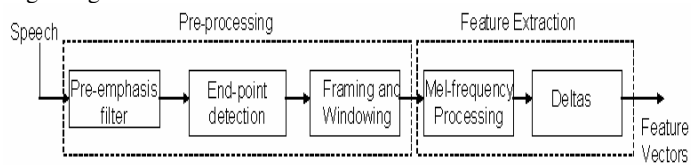

Figure 1: *Speech pre-processing and feature extraction*

Feature extraction extracts relevant features and these features are used to model each speaker [9]. The most popular feature extraction used in speaker recognition is Mel-Frequency Cestrum Coefficients (MFCC). MFCC are spectrum based features and are used in this study since they have shown to give reasonable results. Research also shows that, MFCC feature extraction is more effective when used with its *delta* (DMFCC) and the *delta-delta* (DDMFCC) coefficients [9]. DMFCC and DDMFCC get the dynamics of the speech that do not change. As a result DMFCC and DDMFCC are used throughout this paper.

### 4.2 The classical HMM SID recognizer

The current architecture of automatic HMM based speaker identification system is shown in figure 2. This architecture will be used as the baseline in this study. Practically, speech signal can be corrupted in certain frequency bands. In the conventional SID system this can highly affect the performance of the system, since the entire feature vector is used by the recognizer. Based on these reason the sub-band based SID is expected to perform better than the conventional SID recognizer for "live" testing.

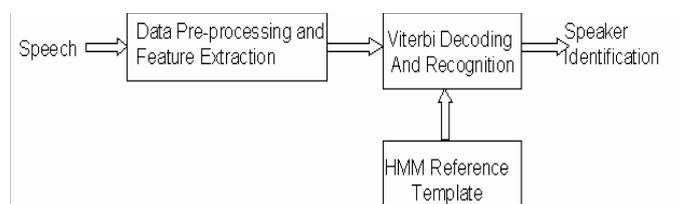

Figure 2: *The classical HMM based SID recognizer*

### 4.3 Sub-band based recognizer

The schematic diagram of the proposed sub-band based SID is presented in figure 3 below. As shown in this figure, the sub-band model begins by dividing the incoming speech spectrum into several sub-bands (1…N). The sub-band HMM classifier estimates the probability for each band. The class conditional log-likelihoods of each sub-band then, serve as input to the merger to give the final speaker ID. The main advantage of this system is that it allows selective deemphasise of unreliable sub-bands.

We treated each sub-band independently and hence feature extraction was performed for each sub-band. The success of this sub-band based recognizer method is highly dependent on the type of merger used. This paper proposes and explores both linear and non-linear merging techniques as shown in figure 3 and further discussed in the next section.

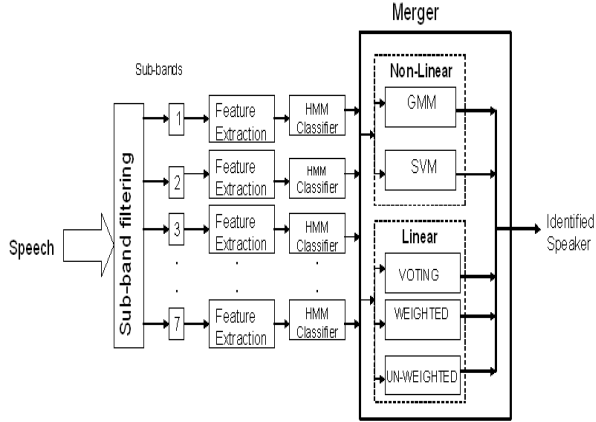

Figure 3: *The sub-band based SID*

*4.3.1 Merging techniques*

There are two types of merging techniques investigated in this paper which are the linear and non-linear mergers. The non-linear mergers are SVM and GMM. These mergers are trained in parallel with the log-likelihoods from each sub-band as input and the corresponding speaker identity as the output. GMM and SVM independently give an identification of a speaker. The linear techniques used are the majority vote rule, the weighted and un-weighted Linear Combination of Log-likelihood Ratios (LCLR) techniques.

*Majority vote*: The first linear technique used is a majority voting rule. In this method, the identified speaker is the one who majority of the sub-bands claim to be the owner of the utterance. Ties are resolved by defining a ordering of each sub band classifier. This method can only be applied for the sub-band based SID with more than 2 bands.

*Weighted and Un-weighted LCLR:* The weighted linear merger combines the log-likelihood of each sub-band using,

$$\hat{p}(X) = \sum_{i=1}^{N} w_i \hat{p}_i(x) \quad (5)$$

where, $x$ is the sub-band data, $X$ is the original input speech signal and $w_i$ are the weights of sub-band $i$. Estimation of the $i^{th}$ weighting factor is computed using,

$$w_i = \frac{IR_i}{\sum_{i=1}^{N} IR_i} \quad (6)$$

where, $IR_i$ is the identification rate of the $i^{th}$ speaker. Equation 6 allocates weights according to the known performance of a sub-band. Thus a sub-band which generally performs better than others is given more weight than the rest of the sub-bands. The un-weighted LCLR on the other hand, is computed almost exactly the same as the un-weighted sum, except it uses equal weights for each band.

*4.3.2 Sub-band selection*

Another challenge other than determining the merging technique is the choice of the number of sub-bands, as well as the frequency spanned by each sub-band. We experimented with the 2, 4 and 7 sub-band systems. The sub-band frequencies were chosen based on the suggestion presented by Allen [10] and are as follows: The two sub-band system had frequencies in the ranges of 0- 1140Hz, 1046-4000Hz. The four sub-band system's frequencies were as follows: 0-765, 400-1640, 1020- 2700, 1860-4000 whereas for the seven sub-band 0-360;330-640, 580-950, 860 -1360, 1265 -1920, 1800-2700, 2515-4000 were used. In addition, the sub-bands were made to overlap to avoid losing features in the edges of the sub-bands.

**4.4 The classical and sub-band based voting system**

Since the main focus of the study is to improve the live testing, an additional approach proposed here is to combine the classical and the sub-band based classifiers using the majority voting. The input in this approach is the vector scores from the individual classifiers and the output is the final speaker's identity. It should be noted that in the occurrence of ties, this is resolved by defining an ordering of the classifiers.

## 5. Confidence and performance measure

**5.1 Performance evaluation of the SID**

Since this is a speaker identification system the main concern is the ability of the system to identify speakers. The performance of the system is measured using the identification rate (IR) given by:

$$IR = \frac{\text{Correctly identified speaker utterance}}{\text{Total speaker utterance}} \times 100\% \quad (7)$$

It should be noted that that this percentage must be low for the impostors and high otherwise. To ensure that this is achieved the confidence of the speaker identification system is measured. The likelihood ratio test is used as the confidence measure in this study and is discussed in the next section.

Once the confidence of the speaker is determined the system may either identify or reject the speaker. The reliability of the system in identifying (i.e. true identification) and rejecting the impostors (i.e. true rejection) is defined by equation 8. This reliability is usually referred to as the decision gap and this is discussed more in the next section.

$$\text{Reliability=Identification rate- True Rejection rate} \quad (8)$$

**5.2 Confidence measure: likelihood ratio test**

Confidence measure addresses the issue of how well a speaker model matches the data, thereby estimating and improving the reliability of the speaker-identification system given by equation 8. In other words, the likelihood ratio test is aimed at determining whether or not the sequence feature vector, *X,* were generated by the family of probability densities of the registered speakers with models $H_{i......N}$. The identification confidence estimation using log-likelihood ratio test is computed by initially defining the log-likelihood ratio (LR) as [11],

$$LR = \frac{P(X \mid H_{claimed\ identity})}{\sum_{i=1}^{N} P(X \mid H_i)} \begin{cases} \geq \tau & accepted \\ > \tau & rejected \end{cases} \quad (9)$$

or,

$$LR = \frac{l_{claimed\ identity} - l_{avg}}{number\ of\ speakers} \quad (10)$$

with,

$$l_{avg} = \frac{1}{N} \sum_{j=1}^{N} l_j \quad (11)$$

where, $l_{claimed\ identity}$ and $l_{avg}$ is the claimed and average log-likelihood scores, respectively.

This quantity is then compared to the threshold $\tau$, were the signal, X, is rejected if the ratio is below this threshold. An important step in the implementation of the likelihood ratio is selecting the optimum decision boundary. The optimum threshold that will increase the identification rate and decreases the impostor identification rate is obtained by using Genetics Algorithm. GA finds this optimum decision boundary by increasing the reliability or the decision gap given by equation 8. A more detailed discussion in using likelihood ratio as confidence measure and determining threshold value has been explained by [11].

## 6. Experimental evaluation and setup

In this section we present, the speech database, along with the experimentation conditions of both baseline and sub-band based recognizers.

### 6.1 Speech database used for experimentation evaluation

Our experiments are based on the database consisting of 20 speakers with 50% females both with different South African accents. Each of the 20 speakers (10 are given access and the rest are impostors) were required to say 'password' 40 times. The speech signals were taken in sessions that were at least a week apart. This is to ensure that all different speaker moods are captures and thereby improving the true identification rate.

Twenty of the forty signals form each of the 10 speakers, were used for training, the other twenty was used for determining the speaker identification rate. On the other hand, all forty signals form the last ten speakers were used to determine impostor rejection rate for both approaches presented here. The speech signals were recorded at a sampling frequency of 16 kHz in laboratory environment.

### 6.2 The baseline recognizer

The first set of experiments addresses the problem of speaker identification using the traditional HMM–based recognizer described in section 3. Using this method as a baseline, exhaustive search was used to find the optimum HMM parameters. It was then found that full covariance matrix and 4 states with 25 Gaussian mixtures, produced the best results for each of the 10 speakers and, was therefore used throughout this study.

### 6.3 The sub-band recognizer

The second set of experiments uses the text-dependent sub-band based system described in section 4. Using this approach, the speech spectrum was divided into 2, 4 and 7 bands using a second order Butterworth filter. As it was for the baseline, the number of states and the Gaussian mixtures used to model the speaker in each sub-band must be selected. This is performed using GA since exhaustive search can be computationally intensive.

We have also conducted a series of experiments using different merging techniques, both linear (voting, weighted and un-weighted linear combination) and non-linear (SVM and GMM) techniques. The last set of experiments investigated the effect of combining the different merging techniques and the baseline classifiers using the voting method.

## 7. Summary of experimentation results

This section presents the experimental results obtained for each recognizer. There are two types of testing that will be used through out this paper which are "live" and "non-live" testing. "Live" testing here is defined as the performance of the system if it is used by the general public. On the other hand, "non-live" testing is performed by splitting the data set into training and testing set. The identification rate obtained when using this testing set is then the "non-live" test results.

### 7.1 Sub-band recognizer

This section investigates the benefits of sub-band recognizer over the baseline recognizer and also determines the benefit of the linear merger over the non-linear merger. Using the speaker database described in section 6, the identification rate obtained for each sub-band of the 2, 4 and 7 sub-band models, is shown in table 1. From this table, it is clear that the accuracy varies with each sub-band. Since the accuracy of each band changes, the final identification rate of the system is expected to be highly depended on the merging technique used. For the non-linear merger, SVM with linear kernels and GMM with 20 spherical Gaussian mixtures produced the best results. The results obtained using both linear and nonlinear merging techniques are shown in table 2.

Table 1: *Identification rate of each band in the [2, 4 7] band model*

|        | Band number (%) |      |      |      |      |    |    |
|--------|-----|------|------|------|------|----|----|
| Model  | 1   | 2    | 3    | 4    | 5    | 6  | 7  |
| 2-band | 88  | 80.5 | -    | -    | -    | -  | -  |
| 4-band | 82.5| 90   | 72.5 | 80.5 | -    | -  | -  |
| 7-band | 82  | 81.5 | 75.5 | 89.5 | 77.5 | 87 | 81 |

Table 2: *Sub-band model identification rate for both linear and non-linear merging techniques*

| Recognizer | Baseline |  |  | 84.5 |  |
|---|---|---|---|---|---|
|  | Linear |  |  | Non-linear |  |
|  | voting | Weighted | Un-Weighted | SVM | GMM |
| 2-band model | 85.5 | 93.5 | 93 | 81 | 99.5 |
| 4-band model | 85.5 | 90.5 | 91 | 82 | 99.5 |
| 7-band model | 89.6 | 89 | 90.5 | 82 | 100 |

This table shows that for voting merger the 7-band model outperforms other band-based models and for this reason 7-band model is used throughout this study for the sub-band based voting merger. Table 2 also illustrates that for both weighted and un-weighted merger 2-band model produced the best results. The non-linear merger on the other hand, seems to significantly outperform the linear mergers for all band-based models. It should be noted that the identification results, shown in table 2 above are for "non-live" testing.

During the study, it was however found that the identification rate of each merging technique behaves quite differently for the "live" testing. Both "live" and "non-live" testing results are shown graphically in figure 4. The figure shows that SVM and GMM merger (i.e. non-linear merger) underperform in "live-testing". This observation opposes the "non-live" testing results shown in table 2. The results in figure 4 show that the identification rate for "live" testing is smaller than the one obtained for the "non-live" testing.

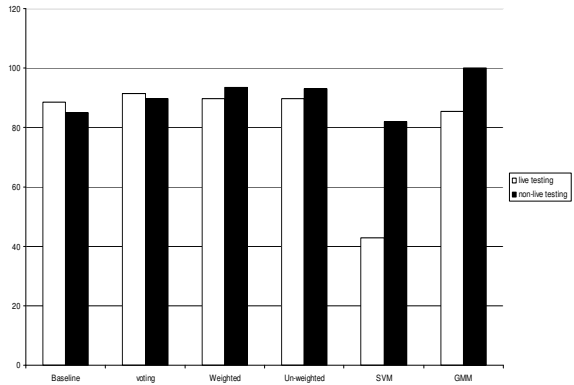

Figure 4: *Live testing and non-live testing for both linear and non-linear merging techniques.*

**7.2 Baseline and sub-band voting system results**

This section presents the results of combining the baseline and sub-band recognizer with linear merging techniques. The voting technique is aimed at improving the "live" testing identification rate. Based on the results obtained in the last section, only linear merging techniques are used in this multiple-classifier voting system. The results obtained for both "live" and "non-live" testing comparing with the baseline for clarity, are shown in figure 5. From figure 5, the multiple-classifier voting system seems to outperform the baseline and sub-band based recognizer for the "live" testing. However, no major improvements are found for the 'non-live" testing.

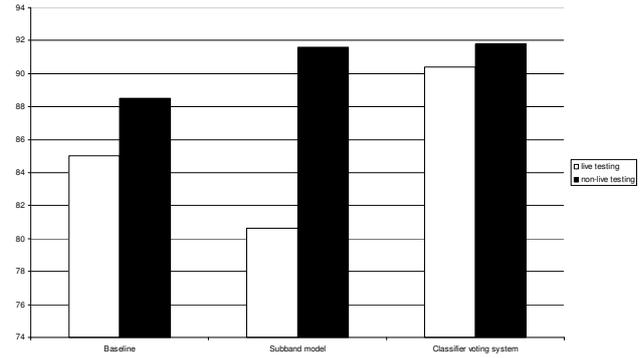

Figure 5: *The identification rate for the multiple–classifier voting system comparing with the baseline and sub-band (weighted) based recognizer*

**7.3 Confidence estimation results**

Using GA algorithm with 25 generations and a population size of 50, the optimum decision gap was found to be 88.5 and 47 for the sub-band and baseline recognizer. The schematic illustration of these results is shown in figure 6. This figure shows that the optimal log-likelihood ratio threshold $\tau$, as 1245 and 1039.8 for the baseline and sub-band method, respectively. The result demonstrates that the sub-band based method does not only produce higher identification rate (see section 2) but also has a higher reliability when compared to the baseline recognizer.

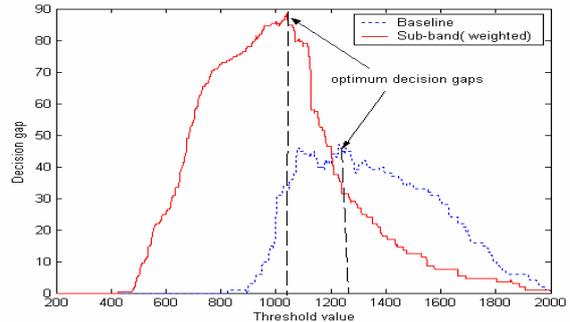

Figure 6: *Threshold versus reliability for both live testing and non-live testing*

The curve in figure 7 was obtained by varying the false acceptance rate of the recognizers and, this is done by changing the decision threshold. The figure shows that decreasing the false acceptance rate increases the false rejection rate as expected. Figure 7 also show that the minimum false identification rate for the sub-band and baseline recognizer is 4% and 35% respectively. On the other hand, the false rejection rate is 28% for the baseline recognizer and 8% for the linear sub-band recognizer.

Figure 8 displays the plot of the probability distribution obtained for the false identification rate of impostors and false rejection of speakers for both baseline and sub-band recognizer. Figure 8 also shows that there is a clear decision boundary for both recognizers and this agrees with the results obtained by GA optimization algorithm.

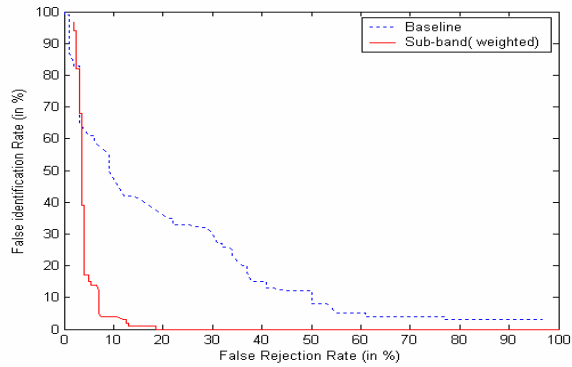

Figure 7: *False acceptance and rejection rate for both live and non-live*

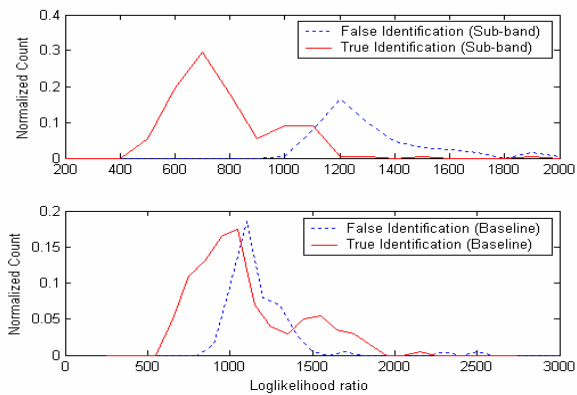

Figure 8: *The schematic illustration of the probability distribution of log-likelihood ratio for true and false identification*

## 8. Discussion and conclusions

Due to dynamic nature of the speech signals, current speaker identification systems produce reasonable results, but still lack the necessary performance if they are to be used the general public. The variability in speech is mainly caused by the length of the vocal track, varying pitch and speaking rate as well as different accents and speaking styles This paper challenged this problem and the results obtained did not only improve the identification rate but also improved the reliability of the SI system.

Results presented in this paper demonstrated that sub-band based speaker recognizer, used with linear merging technique offers enormously improved speaker identification, compared to the wide band system. Tests were performed, comparing the linear and non-linear sub-band recognizer with the conventional speaker identification recognizer for both live and non-live testing. The test revealed that non-linear sub-band based recognizer outperformed the conventional recognizer and the linear sub-band recognizer during non-live testing. The tests further revealed that the linear sub-band based recognizer performed better than both non-linear sub-bands based and conventional the recognizer.

It was further observed that generally the identification rate for live testing was generally smaller than for "non-live" testing. For this reason a voting system was adopted, which improved the 'live testing results by 9.78%. However, no major improvements were observed for the "non-live" testing.

The reliability of the system has also been evaluated and, this was done using the log-likelihood ratio as the identification confidence estimation. Using this log-likelihood ratio, the decision gap size for each method was analyzed, since this is direct measure of the reliability of the identification system. It appears that the decision gap of the linear sub-band based recognizer is approximately twice that of the conventional wide band approach. This enormous improvement in reliability of the system explains why the sub-band approach outperforms the conversional method for "live" testing.

Based on the results obtained there is no doubt that sub-band based recognizer offers a practical solution to the problem of speaker identification. However, the work is still far from being finished. Future work can look at the effect of using different classifier for the sub-band classifier other than HMM. Even though it can be computationally intensive, further work can also look at the effect of using different feature extraction of different size for each band.